\documentclass[10pt,twocolumn,letterpaper]{article}

\usepackage{iccv}
\usepackage{times}
\usepackage{epsfig}
\usepackage{graphicx}
\usepackage{amsmath}
\usepackage{amssymb}

\usepackage{algorithm}
\usepackage{algpseudocode}
\usepackage{amsmath}
\usepackage{graphics}
\usepackage{epsfig}

\usepackage[breaklinks=true,bookmarks=false]{hyperref}

\iccvfinalcopy 

\def\iccvPaperID{****} 

\ificcvfinal\fi

\begin{document}

\title{Video Temporal Relationship Mining for Data-Efficient Person Re-identification}

\author{Siyu Chen\\
Fudan University\\
{\tt\small siyuchen19@fudan.edu.cn}
\and
Dengjie Li\\
Meituan\\
{\tt\small lidengjie@meituan.com}
\and
Lishuai Gao\\
Meituan\\
{\tt\small gaolishuai@meituan.com}
\and
Fan Liang\\
Meituan\\
{\tt\small liangfan02@meituan.com}
\and
Wei Zhang\\
Fudan University\\
{\tt\small weizh@fudan.edu.cn}
\and
Lin Ma\\
Meituan\\
{\tt\small forest.linma@gmail.com}
}

\maketitle
\ificcvfinal\thispagestyle{empty}\fi

\begin{abstract}
   This paper is a technical report to our submission to the ICCV 2021 VIPriors Re-identification Challenge. In order to make full use of the visual inductive priors of the data, we treat the query and gallery images of the same identity as continuous frames in a video sequence. And we propose one novel post-processing strategy for video temporal relationship mining, which not only calculates the distance matrix between query and gallery images, but also the matrix between gallery images. The initial query image is used to retrieve the most similar image from the gallery, then the retrieved image is treated as a new query to retrieve its most similar image from the gallery. By iteratively searching for the closest image, we can \textcolor{red}{achieve accurate image retrieval} and finally obtain a robust retrieval sequence.
\end{abstract}

\section{Introduction}

VIPriors Re-identification Challenge is a subtrack of ICCV 2021 Visual Inductive Priors for Data-Efficient Deep Learning Workshop. The difficulty of this challenge is that no pre-trained weights can be used, and models are to be trained from scratch with limited data. The main objective of the challenge is to obtain the highest Mean Average Precision(mAP) score for person re-identification. The dataset is SynergyReID, which contains images of basketball players and referees that taken from short sequences of basketball games. For the validation and test sets, the query images are persons taken from the first frame, while the gallery images are identities taken from the rest frames. Such dataset characteristics  inspire us to mine the video temporal relationship to help tackling the person re-identification problem. According to common facts, the change between two adjacent frames of a video is slight, so there should be a minimum similarity distance between two adjacent frames. Intuitively, it is more reasonable to iteratively search the nearest neighbors of the current image frame by frame instead of using a single image to retrieve all other video frames. Therefore, we propose a post-processing strategy, namely the video temporal relationship mining. Specifically, the first frame of the video (query image) is used to retrieve the second frame (gallery image). Afterwards,  the second frame is then used to retrieve the next frame. With such an iterative retrieval strategy, the video temporal relationships can be more extensively exploited, resulting in a superior person re-identification performance.

\begin{figure}[t]
\begin{center}
   \includegraphics[width=1\linewidth]{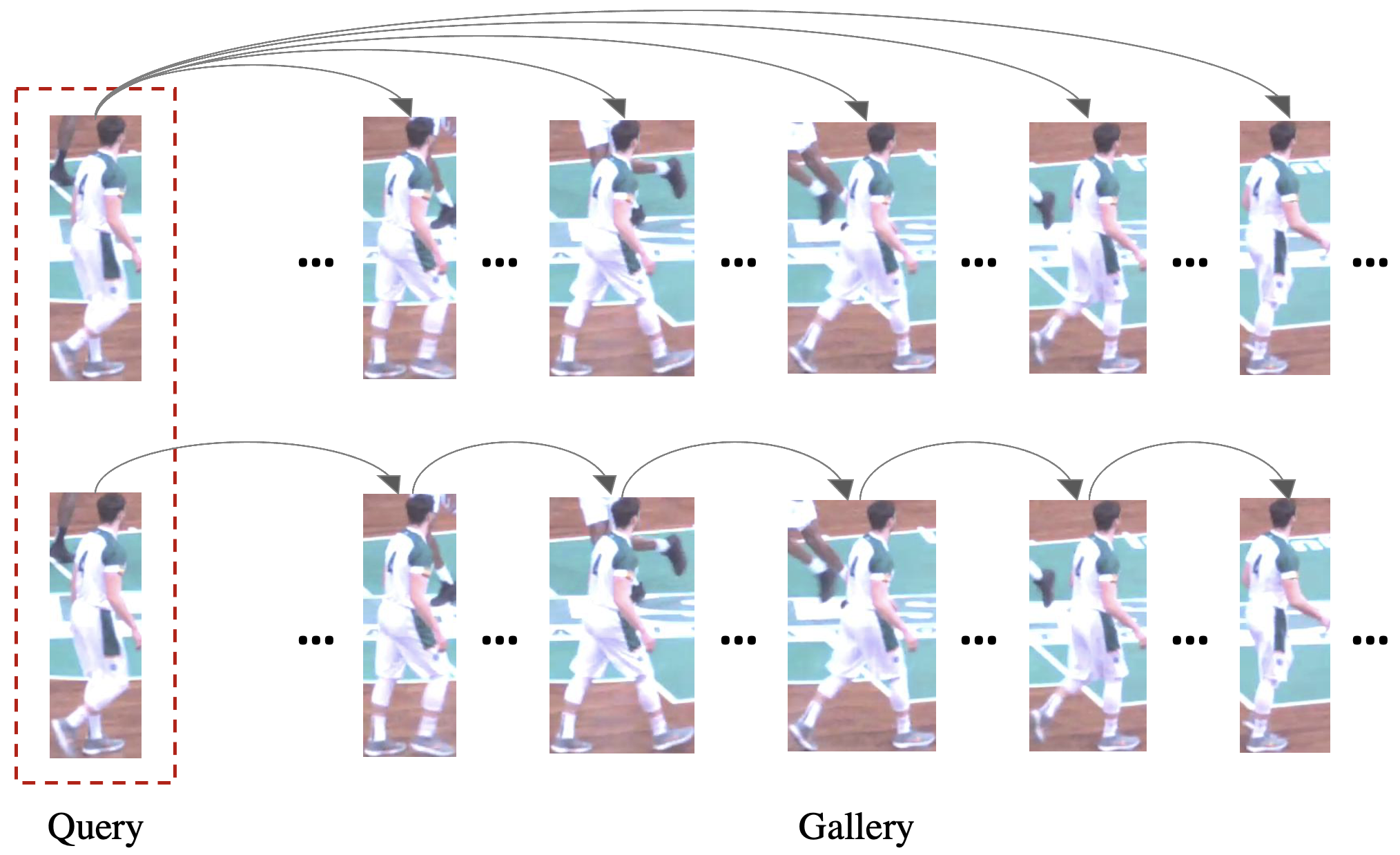}
\end{center}
   \caption{We propose a video temporal relationship mining strategy to iteratively retrieve the closest image from the gallery.}
\label{fig:long}
\label{fig:onecol}
\end{figure}

In addition to using the video temporal relationship mining strategy, we also modify the baseline model to form a new baseline based on the characteristics of the data. Besides, we use model fusion strategy to ensemble the models of different architectures to obtain the best performance. Our work can be summarized as follows.

(1) We carefully select MGN\cite{wang2018learning} as the baseline model. According to the characteristics of the dataset, \textcolor{red}{a fourth branch} is added to the original MGN to make the network learn more discriminative image features.

(2)We propose video temporal relationship mining strategy to make full use of the visual inductive priors of the data. This strategy can greatly improve the performance of person re-identification algorithms. We also discussed different variants of this strategy.

(3)We use model fusion strategies to ensemble models of different architectures. This strategy has further improved our final results.


\begin{figure}[t]
\begin{center}
   \includegraphics[width=0.8\linewidth]{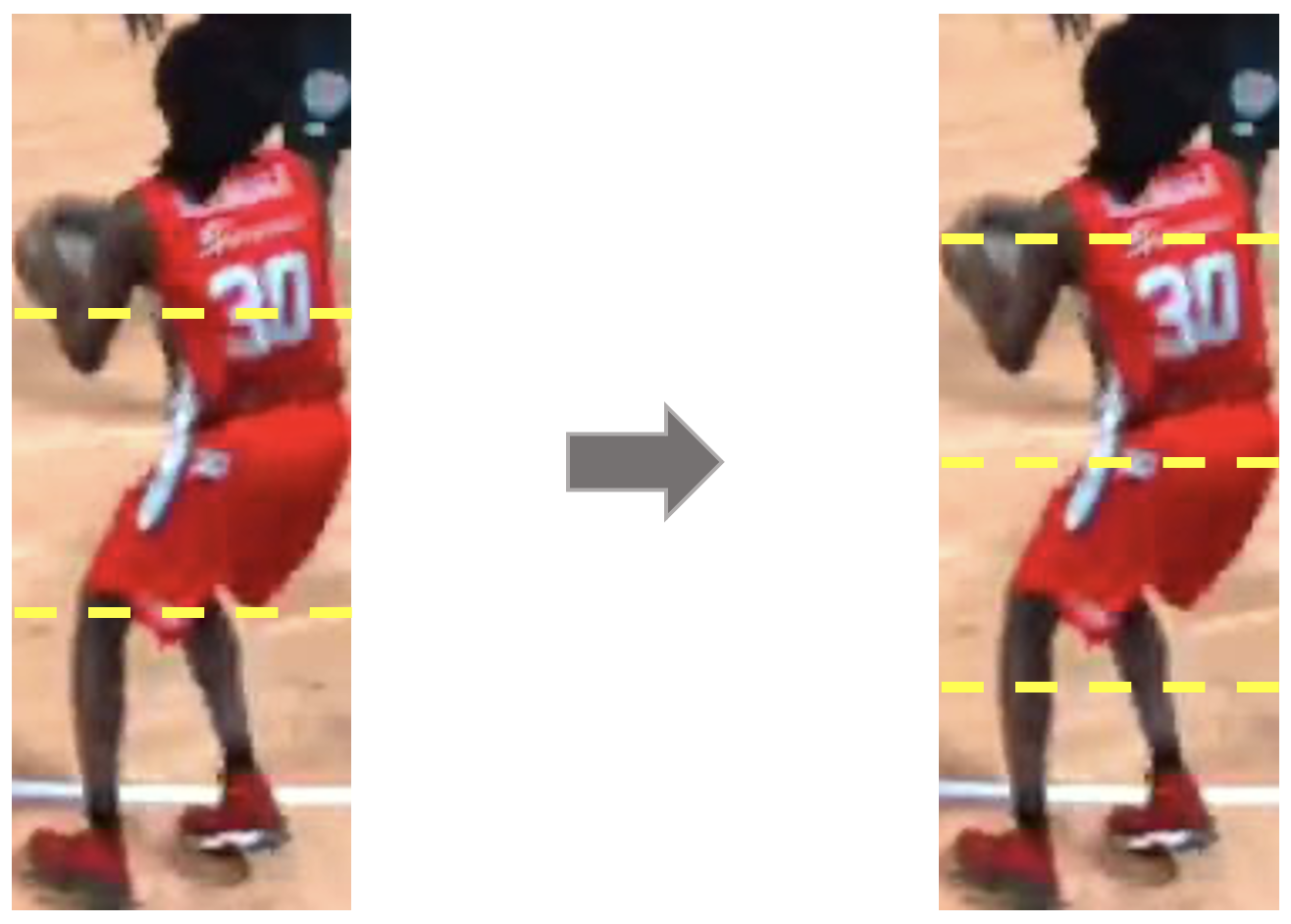}
\end{center}
   \caption{Dividing the image into four parts can avoid destroying the discriminative features(\eg the number on the athlete's shirt) of the image.}
\label{fig:long}
\label{fig:onecol}
\end{figure}

\section{Baseline}

\subsection{Architecture}
We use Multiple Granularity Network(MGN)\cite{wang2018learning} as our baseline model. MGN is a multi-branch deep neural network which composed of a global branch and two local branches. The global branch learns the global feature representations without any feature split operation, and the local branch learns the local representations by splitting image features to several stripes in horizontal orientation. MGN split the image features to two and three parts to form Part-2 Branch and Part-3 Branch respectively. We observed that dividing the feature into 3 parts may damage the discriminative features of the image(\eg the number on the athlete's shirt, see Figure 2.), but only dividing the feature into 2 parts lacks sufficient granularity. Therefore, we keep the original structure of MGN and simply add a fourth branch to form our new baseline model. Table 1 shows the performance improvement after adding Part-4 Branch.

\subsection{Implementation Details}
In order to achieve rapid implementation, we use the FastReID\cite{he2020fastreid} open source framework for all experiments. Our backbone is ResNet-50\cite{he2016deep} with IBN block\cite{pan2018IBN-Net}. We replace BN\cite{ioffe2015batch} with Synchronized BN (SyncBN)\cite{peng2018megdet} for normalization and choose cross-entropy loss and triplet loss to train the network. The weight decay factor for L2 regularization is set to 5e-4. For optimization, the optimizer is Adam\cite{kingma2014adam} and the initial learning rate is set to 3.5e-4. The total training process lasts for 120 epochs. We decrease the learning rate to 3.5e-5 and 3.5e-6 after training for 40 and 90 epochs. We use warning up strategy for the first 2000 iterations and freeze the weights of backbone and branch 1-4 for the first 1000 iterations. We use 4 NVIDIA Tesla V100 GPUs and the total batch size is set to 128. Each GPU contains 4 instance. We follow BoT\cite{luo2019bag} to use Last Stride and BNNeck tricks. During training and testing, images are resized to 384$\times$128. For data augmentation, we use random erase augmentation, random horizontal flipping and padding. Besides, we also deploy reRank\cite{zhong2017re} and so-called 6$\times$ Schedule as proposed in \cite{he2019rethinking}. For 6$\times$ Schedule, we simply expand total training epoch from 120 to 720 and decrease the learning rate by a scale of 1/10 at epoch 240 and 480, respectively. Table 1 shows the performance of our new baseline model after adding Part-4 Branch and using 6$\times$ Schedule.

\begin{table}
\begin{center}
\begin{tabular}{|l|c|}
\hline
~~~Strategies & mAP / +RK \\
\hline\hline
~~~MGN\cite{wang2018learning} & 84.3 / 92.9 \\
+ Part-4 Branch & 86.3 / \textbf{94.6} \\
+ 6$\times$ Schedule & \textbf{88.0} / 94.3\\
\hline
\end{tabular}
\end{center}
\caption{Our baseline results on SynergyReID validation set. “+RK” means use reRank.}
\end{table}
\section{Methods}

\begin{figure*}
\begin{center}
    \includegraphics[width=1\linewidth]{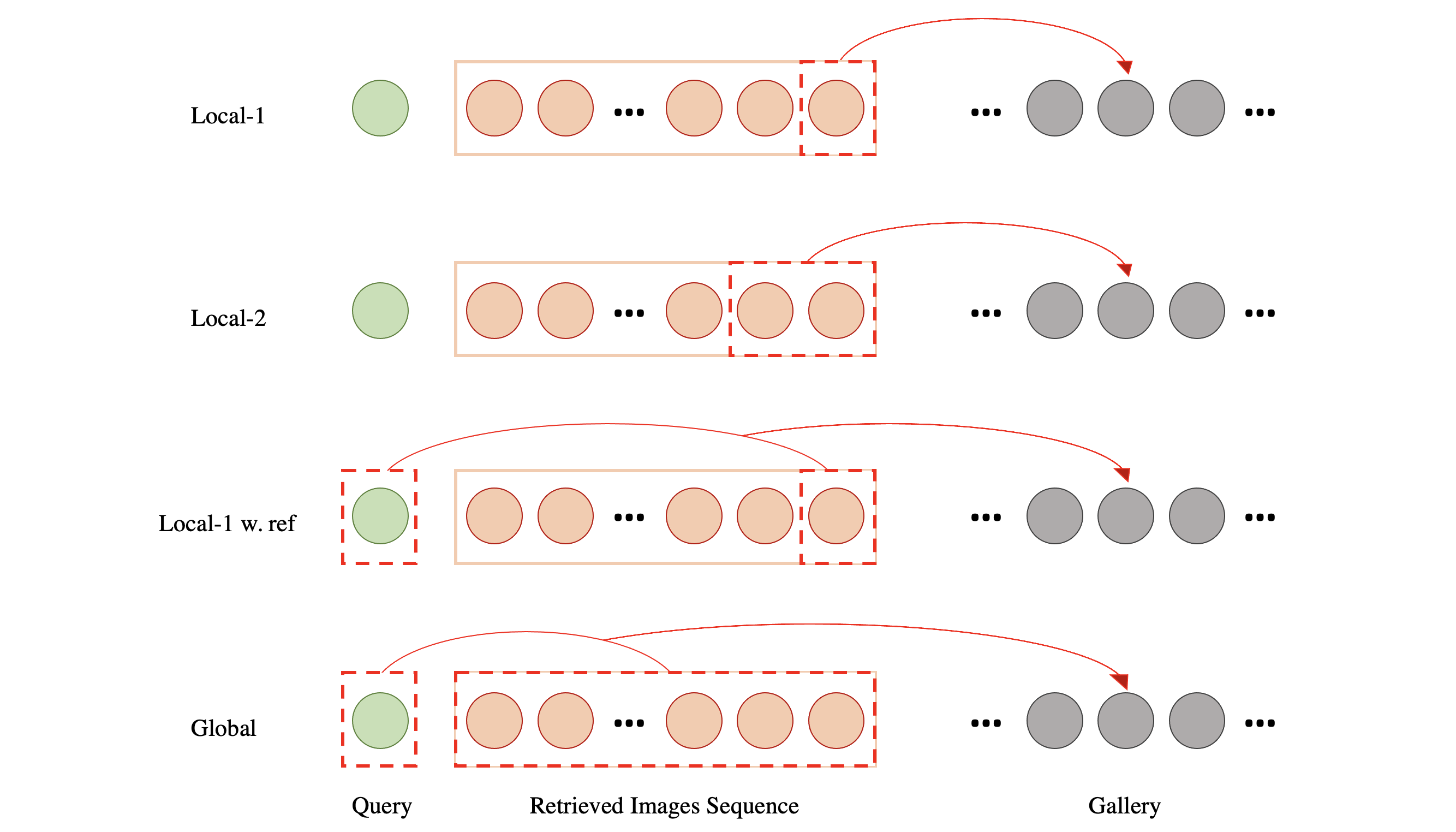}
\end{center}
   \caption{Variations of our proposed video temporal relationship mining strategy. “Local” means to use a part of the most recently retrieved images for the next retrieval. “-1” means the images window size is 1, etc. “w. ref” indicates to retain the original query image as a reference for retrieval. “Glocal” means use all retrieved images for the next retrieval.}
\label{fig:short}
\end{figure*}
\subsection{Video Temporal Relationship Mining}
After training, we can use the features extracted by the neural network to perform image-to-image similarity matching according to a specific distance metric. Generally speaking, the image sequence retrieved by the query is a sorted list based on the distance between this query and gallery. When person with occlusion or similar appearance appear, the retrieval results will be relatively poor. Inspired by the video vision task, we modeling the person re-identification task as a problem of iteratively retrieving the nearest frame of the current frame. Our strategy can deal with senarios such as occlusion and appearance changes well. Furthermore, not only the person characters but also the images' background can be very helpful in discriminating identities.

Specifically, let $\mathcal{Q}$ denote the query image set, and $\mathcal{G}$ denote the gallery image set. $\textit{\textbf{M}}^{(\mathcal{Q},\mathcal{G})}$ is the distance matrix between query images and gallery images, where each element $m_{ij}$ of the matrix represents the similarity distance between the $\textit{i}$-th query image $\textit{q}_{i}\in\mathcal{Q}$ and the $\textit{j}$-th gallery image $\textit{g}_{j}\in\mathcal{G}$. Let $\mathcal{R}$ denote the retrieved image list and $\mathcal{R}_{i}\subset\mathcal{R}$ is the sequence that retrieved by $\textit{q}_{i}$. At the beginning of retrieval, we regard query $\textit{q}_{i}$ as the first frame of a video, and retrieve the nearest gallery image $\textit{g}$ based on matrix $\textit{\textbf{M}}^{(\mathcal{Q},\mathcal{G})}$. We delete the retrieved image $\textit{g}$ from $\mathcal{G}$ and add it to retrieved image sequence $\mathcal{R}_{i}$, then we denoted it by $\textit{r}_{i1}\in\mathcal{R}_{i}$. For the next retrieval, we use $\textit{r}_{i1}$ to retrieve its nearest image in gallery. Thus, we calculate distance matrix $\textit{\textbf{M}}^{(\mathcal{G},\mathcal{G})}$, where each element $\textit{m}_{ij}$ of the matrix represents the similarity distance between $\textit{g}_{i}\in\mathcal{G}$ and $\textit{g}_{j}\in\mathcal{G}$. The retrieved image by $\textit{r}_{i1}$ is deleted from $\mathcal{G}$ and added to $\mathcal{R}_{i}$ as $\textit{r}_{i2}$. By iteratively retrieve the nearest image from gallery, we can obtain the final retrieved image sequence $\mathcal{R}_{i}$. The process of video temporal relationship mining strategy is shown in Algorithm 1. Our final output after using this strategy is an index matrix based on $\mathcal{R}$.

\begin{algorithm}
    \caption{Video Temporal Relationship Mining(VTRM)}
    \label{euclid}
    \begin{algorithmic}[1] 
        \Procedure{VTRM}{$\textit{\textbf{M}}^{(\mathcal{Q},\mathcal{G})},\textit{\textbf{M}}^{(\mathcal{G},\mathcal{G})}$}
            \State $\textit{\textbf{M}}^{(\mathcal{Q},\mathcal{G})}$: Distance matrix between query and gallery images with shape $m\times n$
            \State $\textit{\textbf{M}}^{(\mathcal{G},\mathcal{G})}$: Distance matrix between gallery images with shape $n\times n$
            \State $\mathcal{Q}$: Query image set
            \State $\mathcal{G}$: Gallery image set
            \State $\mathcal{R}$: Retrieved image list which initialized with $\emptyset$
            \For{each $i\in [1,m]$} 
                \State initialize a set $\mathcal{G}^{i}$ with gallery images;
                \State $\mathcal{R}_{i} = \emptyset$;
                \State $\textit{g} \gets$ image of minimum dist with $\mathcal{Q}_{i}$ based on $\textit{\textbf{M}}^{(\mathcal{Q},\mathcal{G})}$;
                \State $\mathcal{R}_{i} = \mathcal{R}_{i} + \textit{g}$;
                \State $\mathcal{G}^{i} = \mathcal{G}^{i} - \textit{g}$;
                \For{each $j\in [2,n]$}
                    \State $\textit{g} \gets$ image of minimum dist with \textit{g} in $\mathcal{G}^{i}$ based on $\textit{\textbf{M}}^{(\mathcal{G},\mathcal{G})}$;
                    \State $\mathcal{R}_{i} = \mathcal{R}_{i} + \textit{g}$;
                    \State $\mathcal{G}^{i} = \mathcal{G}^{i} - \textit{g}$;
                \EndFor 
                \State $\mathcal{R} = \mathcal{R} + \mathcal{R}_{i}$
            \EndFor
            \State \textbf{return} $\mathcal{R}$
        \EndProcedure
    \end{algorithmic}
\end{algorithm}

\begin{algorithm}
    \caption{Model Fusion}
    \label{euclid}
    \begin{algorithmic}[1] 
        \Procedure{ModelFusion}{$\mathcal{S},\mathcal{M}$}
            \State $\mathcal{S}$: Set of $K$ retrieval lists $\mathcal{R}$
            \State $\mathcal{M}$: Set of $K$ distance matrices $\textit{\textbf{M}}^{(\mathcal{Q},\mathcal{G})}$, each matrix has a shape of $m\times n$
            \State $\mathcal{C}$: Candidates set
            \State $\mathcal{V}$: Votes of each candidates
            \State $\mathcal{O}$: Output of model fusion
            \For{each $i\in [1,m]$} 
                \State $\mathcal{O}_{i} = \emptyset$;
                \For{each $j\in [1,n]$}
                    \State $\mathcal{C} = \emptyset$;
                    \For{each $l\in [1,K]$} 
                        \State $c \gets$ the first image of $\mathcal{R}_{i}^{l}$ and not in $\mathcal{O}_{i}$
                        \If {$c \in\mathcal{C}$}
                            \State $v_{c} = v_{c}+1$;
                        \Else
                            \State $\mathcal{C} = \mathcal{C}+c$
                            \State $v_{c} = 1$;
                        \EndIf
                    \EndFor
                    \State $\mathcal{C} \gets$ select the highest voting candidates based on $\mathcal{V}$
                    \State $k=len(\mathcal{C})$
                    \If{$k==1$}
                        \State $c \gets$ the image in $\mathcal{C}$
                    \Else
                        \State $c \gets$ select the nearest image in $\mathcal{C}$ based on $\mathcal{M}$
                    \EndIf
                    \State $\mathcal{O}_{i} = \mathcal{O}_{i} + c$
                \EndFor
                \State $\mathcal{O} = \mathcal{O} + \mathcal{O}_{i}$
            \EndFor
            \State \textbf{return} $\mathcal{O}$
        \EndProcedure
    \end{algorithmic}
\end{algorithm}

\subsection{Variants}

\begin{table}
\begin{center}
\begin{tabular}{|l|c|}
\hline
Variants & mAP(+RK) \\
\hline\hline
Baseline & 94.3 \\
Local-1~~/ w. ref & 97.1 / 95.7 \\
Local-2~~/ w. ref &\textbf{97.3} / 95.8 \\
Local-3~~/ w. ref & 95.7 / 95.8 \\
Local-4~~/ w. ref & 95.8 / 96.0 \\
Local-5~~/ w. ref & 96.0 / 96.0 \\
Local-10/ w. ref & 95.5 / 96.1 \\
Global & 93.9 \\
\hline
\end{tabular}
\end{center}
\caption{Performance comparison of different variants on SynergyReID validation set. All reported results have used the reRank strategy. “Baseline” represents the model described in section 2.2. “Local-1” indicates to use local retrieved images to retrieve next image with the window size set to 1, etc. “w. ref” indicates to retain the original query image as a reference for retrieval. “Global” indicates to use all retrieved images for the next retrieval.}
\end{table}

In this section, we discuss several variants based on the strategy we proposed in the previous section. We observe that the retrieval process based on a certain image actually sets a retrieval image window, which starts from the end of the retrieved image sequence and ends according to the window size. We name this retrieval process “\textbf{Local}” retrieval, which corresponds to “\textbf{Global}” retrieval, that is, query  and all retrieved images will participate in the next retrieval process. For local retrieval, there are "\textbf{Local-N}" variants according to the window size $\textbf{N}$. The video temporal relationship mining strategy described in the previous section is "Local-1". In addition, according to whether to keep the original query as a reference when doing retrieval, local retrieval can be divided into "Local" and "Local w.ref". The illustration is shown in Figure 3.

Table 2 shows the results of using different video temporal relationship mining variants. The baseline model have described in section 2.2. Below we analyze the reasons for the difference in performance of these variants.

\noindent\textbf{Local-N}~~~In our experiments, the performance of “Local-2” is the best. We have tried other models with different architecture, and found that “Local-1” or “Local-2” was always among the best results. We guess that there are two reasons for this phenomenon. First, this is related to the frame rate of the video (or the dissimilarity of the gallery images). Second, the larger the retrieval window, the more likely it is to introduce erroneous results, which will pollute the search window. For SynergyReID, the frame rate of original video clips is slow, so “Local-1” or “Local-2” always achieve best results.

\noindent\textbf{Local-N w.ref}~~~For all the models that using “w.ref” strategy, their performance fluctuations are quite small(within 0.4\% mAP). We speculate that the retrieval process that retains the query as a reference is more stable. For SynergyReID, the query image is the first frame of a video clip. If we choose a middle frame of the video as the query, the “Local-N w.ref” strategy may perform better than “Local-N”.

\noindent\textbf{Global}~~~Global strategy perform worse in all of our experiments. We speculate that this strategy makes the retrieval sequence extremely easy to be contaminated. Imagine that in a retrieval process, a image that originally belonged to another identity was incorrectly retrieved, then in the next retrieval, other images of this wrong identity will be easily retrieved, and then continue to contaminate the retrieval sequence.

\subsection{Model Fusion}

\begin{figure*}
\begin{center}
    \includegraphics[width=0.9\linewidth]{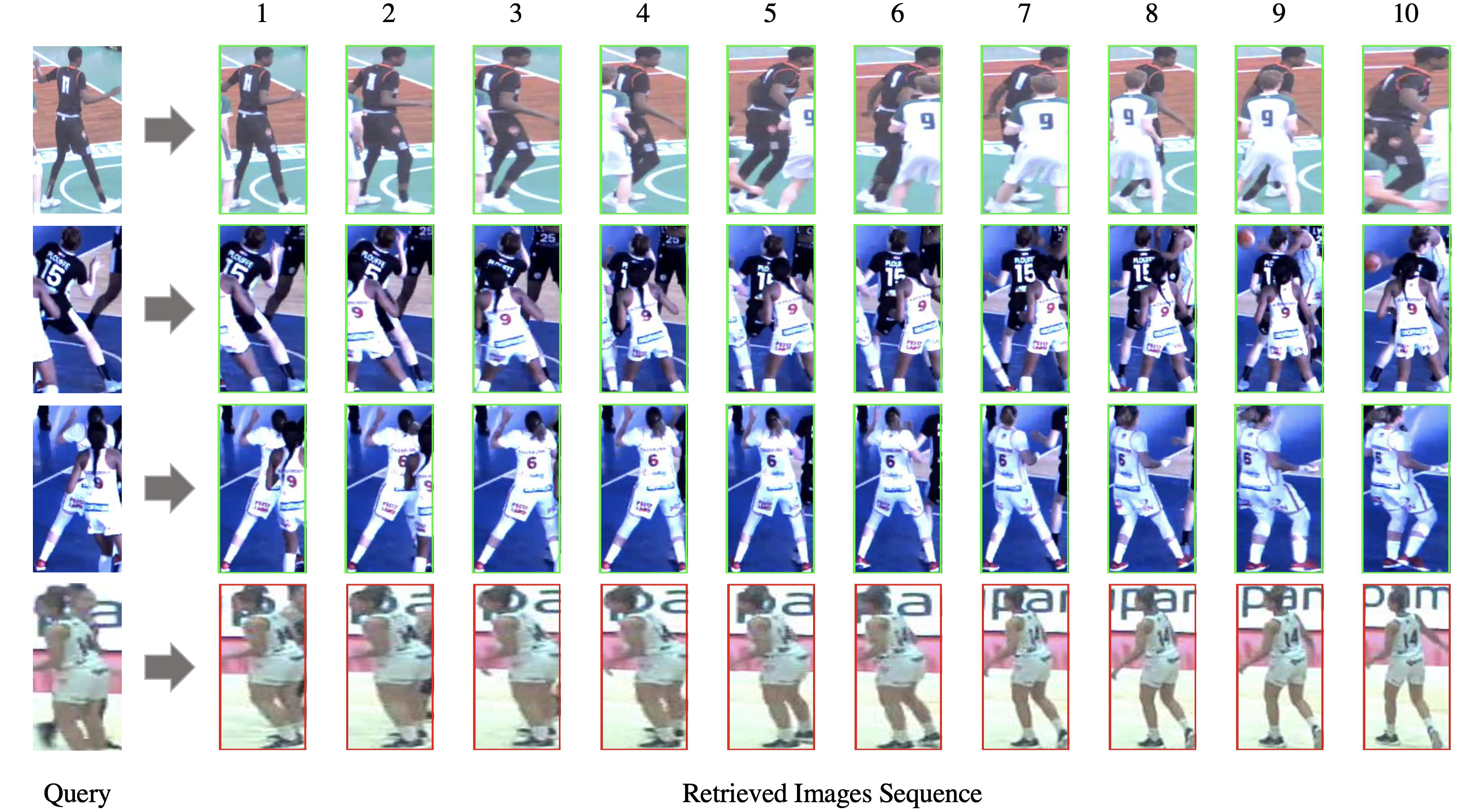}
\end{center}
   \caption{Visualization of our results on SynergyReID validation set. The first three rows shows the case with occlusion and appearance changes. The last row shows the failure case. An image with a green border indicates a correct retrieval, while an image with a red border is not.}
\label{fig:short}
\end{figure*}

\begin{table}
\begin{center}
\begin{tabular}{|l|c|}
\hline
~~~Models & mAP(+RK) \\
\hline\hline
~~~R50-ibn ~~1$\times$ Schedule Local-2 & 96.7 \\
~~~R50-ibn ~~6$\times$ Schedule Local-2 & 97.3 \\
seR50-ibn ~~6$\times$ Schedule Local-2 & 95.5 \\
~~~R101-ibn 6$\times$ Schedule Local-2 & 95.1 \\
seR101-ibn 6$\times$ Schedule Local-2 & 96.6 \\
\hline
5-Model Ensemble & \textbf{98.0} \\
\hline
\end{tabular}
\end{center}
\caption{Results on SynergyReID validation set that using model fusion strategy to ensemble 5 models with different architecture or training scheme.}
\end{table}

We trained four models of different architectures: ResNet50-ibn, ResNet50-ibn with SE module\cite{hu2018senet}, ResNet101-ibn and ResNet101-ibn with SE module. They are all trained with 6$\times$ Schedule. We use model fusion strategy to ensemble them. In order to enrich the fusion results, we added a ResNet50-ibn model that trained with 1$\times$ Schedule.

For model fusion strategy, we first adopt the voting strategy to select the image with the highest number of votes. When there is a tie, we select the image with the closest distance to the original query from the tie candidates as the final result. Assuming that $K$ models need to be fused. Let $\mathcal{R}^{l}$ denotes the retrieved image list of model $l$, and $\textit{\textbf{M}}^{(\mathcal{Q},\mathcal{G})}_{l}$ denotes the query-gallery distance matrix of model $l$, where $l=\{1,2,…,K\}$. Use $\mathcal{O}$ to represent the output of model fusion and $\mathcal{O}_{i}\subset\mathcal{O}$ is the fused retrieved image sequence of query image $\textit{q}_{i}$. Each element $o_{ij}\in\mathcal{O}_{i}$ is obtain by the model fusion strategy. First, traverse the results $\mathcal{R}_{i}$ of all models and search the first image $\textit{r}\notin\mathcal{O}_{i}$ to be a candidate. Then select the highest voting image to be the fusion reslut $o_{ij}$. When there is a tie, every die candidates need to compare its distance to query image. Suppose there are $k$ candidates with tie votes, whose index are $I_{1}$, $I_{2}$, …, $I_{k}$, and their corresponding models are $l_{1}$, $l_{2}$, …, $l_{k}$, respectively. We compare the value of $m_{l_{1}-iI_{i}}$, $m_{l_{2}-iI_{2}}$, …, $m_{l_{k}-iI_{k}}$ , where each $m_{l-iI}$ denotes the element $(i, I)$ in $\textit{\textbf{M}}^{(\mathcal{Q},\mathcal{G})}_{l}$. Then we choose the candidate with the minimum distance as the fusion reslut $o_{ij}$. The process of model fusion is shown in Algorithm 2. After fusing five models, our result reach 98.0\% mAP on SynergyReID validation set. Results are shown in Table 3.

\begin{table}
\begin{center}
\begin{tabular}{|l|c|}
\hline
~~~Strategies & mAP(+RK) \\
\hline\hline
~~~MGN & 90.4 \\
+ Part-4 Branch & 91.8 \\
+ Local-1 & 93.0 \\
+ 6$\times$ Schedule  & 94.0 \\
+ 5-Model Ensemble & \textbf{96.4} \\
\hline
\end{tabular}
\end{center}
\caption{Results on SynergyReID test set using all of the mentioned strategies.}
\end{table}

\subsection{Final Results}

\begin{figure*}
\begin{center}
    \includegraphics[width=1\linewidth]{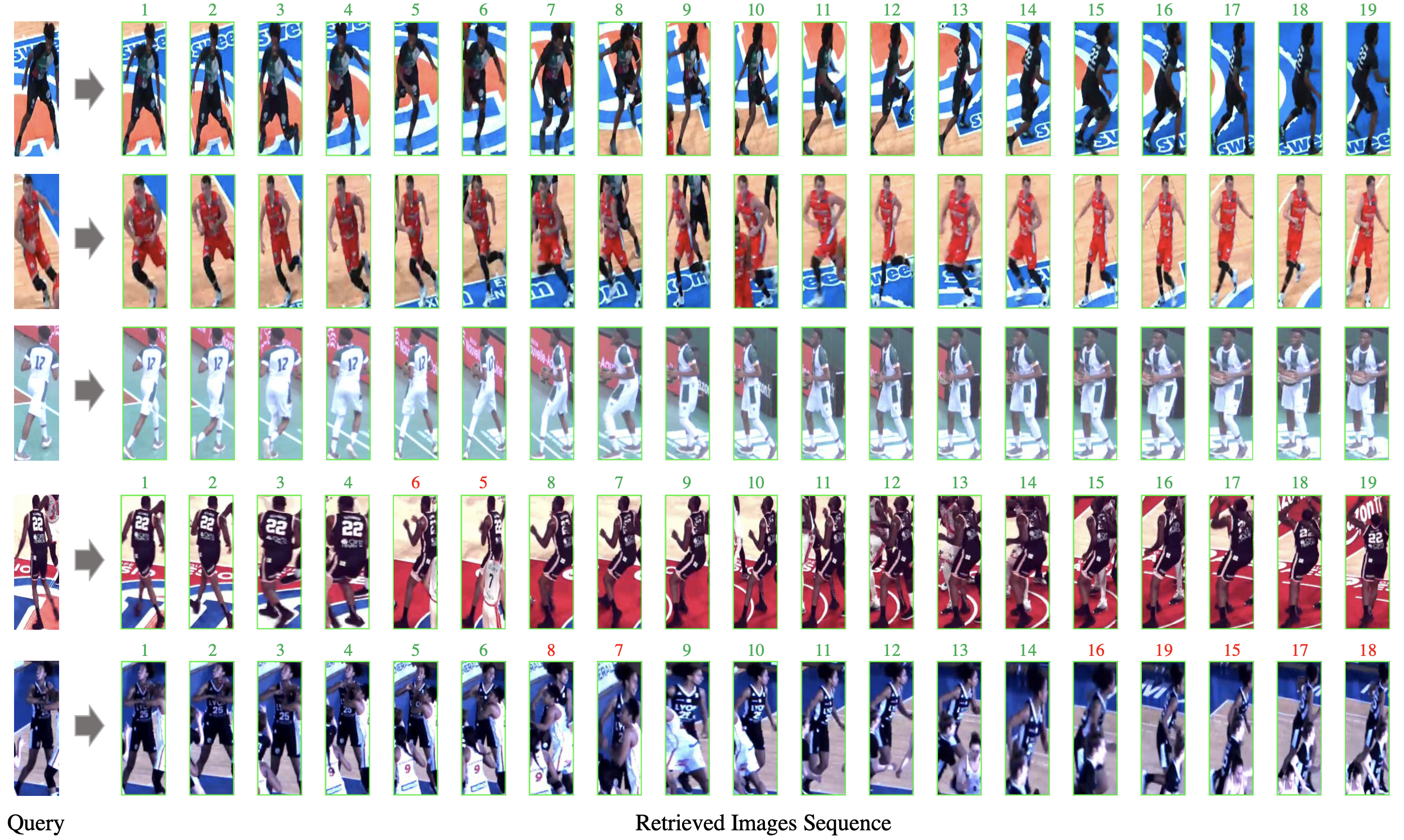}
\end{center}
   \caption{Comparison of video temporal relationship mining results. The number above image indicates the real frame order of the image in video. The green indicates that the order of the image we retrieved is consistent with the real video frame order, while the red indicates an error.}
\label{fig:short}
\end{figure*}

Table 4 shows our results on SynergyReID test set after using all mentioned strategies. The backbone of reported model is ResNet50-ibn. MGN with Part-4 Branch reaches 91.8\% mAP. When using the proposed video temporal relationship mining strategy with “Local-1”, the score rises by 1.2\%. We further improve the score to 94\% after deploy 6$\times$ Schedule. The final results reaches 96.4\% when employ our model fusion strategy. The architecture of these five fusion models have been introduced in section 3.3.

Visualization of our results on SynergyReID validation set is shown in Figure 4. The first three rows shows the case with occlusion and appearance changes. In the first and second lines of the example, the occlusion occurs in the gallery images, while the third line is that the query itself is occluded. Nonetheless, our proposed strategies can handle these scenarios well. The last row shows the failure case. There is no correct retrieval in the first ten retrieved image sequence. But this is mainly because the identity in the query image is very severely occluded.

Figure 5 compares the proposed video temporal relationship mining results with the real video temporal sequence order. All five examples have achieved 100\% mAP. In the first three rows, the order of the images we retrieved is exactly the same as the real video frame order. We can observe that the background information of the image can also help our method to achieve accurate retrieval, even if the query image looks very different from the gallery images(row 3). In some adjacent frames of the fourth and fifth rows, the order we retrieved is wrong with the real video frame order. But it can be seen that these frames in the wrong order are actually very similar. Even some wrong case could be mistaken for the real order(frame 7 and 8 of the last line). Therefore, the strategies we proposed is not to deliberately search the true next frame of the current image, but to find the image that looks most similar to the current image.

\section{Conclusion}
In this report, we propose a strategy of video temporal relationship mining to solve person re-identification task. In the process of participating in the competition, we also modified the baseline model and used the model fusion strategy. With no pre-trained weights and very limited data, we finally achieved 96.4\% mAP on the SynergyReID test set, which ranking second in the ICCV 2021 VIPriors Re-identification Challenge.



\end{document}